\documentclass{article}

\usepackage{microtype}
\usepackage{graphicx}
\usepackage{subfigure}
\usepackage{booktabs} % for professional tables

\usepackage[hidelinks]{hyperref}

\usepackage[accepted]{icml2021}

\icmltitlerunning{Isometric Gaussian Process Latent Variable Model}

\usepackage[utf8]{inputenc} % allow utf-8 input
\usepackage[T1]{fontenc}    % use 8-bit T1 fonts
\usepackage{amsfonts}       % blackboard math symbols
\usepackage{nicefrac}       % compact symbols for 1/2, etc.
\usepackage{bm}
\usepackage{amsmath}
\usepackage{amsthm}
\usepackage{amssymb}
\usepackage{xspace}
\usepackage{extarrows}
\usepackage{xfrac}
\usepackage{multirow}
\usepackage{enumitem}
\usepackage{tikz}
\usetikzlibrary{arrows.meta}
\newtheorem{definition}{Definition}

\usepackage{amssymb}% http://ctan.org/pkg/amssymb
\usepackage{pifont}% http://ctan.org/pkg/pifont
\newcommand{\cmark}{\ding{51}}%
\newcommand{\xmark}{\ding{55}}%

\renewcommand{\mathbf}{\boldsymbol} % this ensures that Greek alphabets are also bold (it will slant, though)

\begin{document}
\twocolumn[
\icmltitle{Isometric Gaussian Process Latent Variable Model \\
           for Dissimilarity Data}




\begin{icmlauthorlist}
\icmlauthor{Martin Jørgensen}{ox}
\icmlauthor{Søren Hauberg}{dtu}
\end{icmlauthorlist}

\icmlaffiliation{ox}{Department of Engineering Science, University of Oxford}
\icmlaffiliation{dtu}{Department of Mathematics and Computer Science, Technical University of Denmark}

\icmlcorrespondingauthor{Martin Jørgensen}{martinj@robots.ox.ac.uk}
\icmlcorrespondingauthor{Søren Hauberg}{sohau@dtu.dk}

\icmlkeywords{Machine Learning, ICML}

\vskip 0.3in
]



\printAffiliationsAndNotice{}  % leave blank if no need to mention equal contribution

\begin{abstract}
We present a probabilistic model where the latent variable respects both the distances and the topology of the modeled data. The model leverages the Riemannian geometry of the generated manifold to endow the latent space with a well-defined stochastic distance measure, which is modeled locally as Nakagami distributions. These stochastic distances are sought to be as similar as possible to observed distances along a neighborhood graph through a censoring process. The model is inferred by variational inference based on observations of pairwise distances. We demonstrate how the new model can encode invariances in the learned manifolds.
\end{abstract}

\section{Introduction}
\emph{Dimensionality reduction} aims to compress data to a lower dimensional representation while preserving the underlying signal and suppressing noise. Contemporary nonlinear methods mostly call upon the \emph{manifold assumption} \citep{bengio:tpami:2013} stating that the observed data is distributed near a low-dimensional manifold embedded in the observation space. Beyond this unifying assumption, methods often differ by focusing on one of three key properties (Table~\ref{tab:methods}).

% Dimensionality reduction is a scientific discipline which primary target is extracting useful information from high-dimensional data.
% The curse of dimensionality, otherwise known as the empty-space phenomenon, is a burden that methods
% that we know to work in the conceiveable spatial dimension, say two and three, do not generalize to higher dimensions.
% The manifold assumption, studied in .. , informs that data in these high dimensional spaces tend to live on manifolds,
% that can be described by fewer variables than what is measured. Hence non-linear dimensionality reduction can help
% us understand data residing in high dimensions. 
%
% In this view, dimensionality reduction has a place within machine learning - more precisely in unsupervised learning,
% where data is collected, but unlabelled. 
% The most famous method in dimensionality reduction is the linear approach \emph{Principle Component Analysis} (PCA). 
	
% Known methods in dimensionality reduction can coarsely be divided into three categories: topological preservation, distance preservation and \emph{generative} approaches; the two first of which can also be categorized as \emph{discriminative}.

\begin{table}
%\vspace{-2mm}
%\centering
%\scalebox{0.9}{%
\resizebox{\columnwidth}{!}{\begin{tabular}{l|ccc}
                & Probabilistic  & Topology  & Distance \\ \hline
PCA             & (\cmark)    & \xmark    & (\cmark) \\
MDS             & \xmark      & \xmark    & \cmark   \\
IsoMap          & \xmark      & (\xmark)  & \cmark   \\
t-SNE           & \xmark      & (\cmark)  & \cmark   \\
UMAP            & \xmark      & \cmark    & \cmark   \\
GPLVM           & \cmark      & \xmark    & \xmark   \\
Iso-GPLVM (our) & \cmark      & \cmark    & \cmark   
\end{tabular}
}
\caption{A list of common dimensionality reduction methods and coarse overview of their features.}
\label{tab:methods}
\end{table}

\paragraph{Topology preservation.}
A \emph{topological space} is a set of points whose \emph{connectivity} is invariant to continuous deformations. For finite data, connectivity is commonly interpreted as a clustering structure, such that topology preserving methods do not form new clusters or break apart existing ones. For visualization purposes, the \emph{uniform manifold approximation projection (UMAP)} \citep{mcinnes2018umap} appears to be the current state-of-the-art within this domain.

\paragraph{Distance preservation.}
Methods designed to find low-dimensional representation with pairwise distances that are similar to those of the observed data may generally be viewed as a variant of \emph{multi-dimensional scaling (MDS)} \citep{ripley2007pattern}. Usually, this is achieved by a direct minimization of the \emph{stress} defined as
\begin{equation}\label{eq:stress}
    \text{stress} = \sum_{i<j\leq N} (d_{ij}-\|\mathbf{z}_i-\mathbf{z}_j\|)^2,
\end{equation}
where $d_{ij}$ are the \emph{dissimilarity} (or \emph{distance}) of two data points $\mathbf{x}_i$ and $\mathbf{x}_j$, and $\mathcal{Z}=\{\mathbf{z}_i\}_{i=1}^N$ denote the low-dimensional representation in $\mathbb{R}^q$. 

More advanced methods have been built on top of this idea. In particular, \emph{IsoMap} \citep{tenenbaum:global:2000} computes $d_{ij}$ along a neighborhood graph using Dijkstra's algorithm. This bears some resemblance to \emph{t-SNE} \citep{maaten2008visualizing} that uses the Kullback-Leibler divergence to match distribution in low-dimensional Euclidean spaces with the data in high dimensions.

\paragraph{Probabilistic models.}
A common trait for the mentioned methods is that they learn features in a mapping from high-dimensions to low, but not the reverse. This makes the methods mostly useful for visualization.
\emph{Generative models} \citep{kingma:iclr:2014, rezende:icml:2014, lawrence2005probabilistic, goodfellow:nips:2014, rezende2015variational} allow us to make new samples in high-dimensional space. Of particular relevance to us, is the \emph{Gaussian process latent variable model (GP-LVM)} \citep{lawrence2005probabilistic, titsias2010bayesian} which learns a stochastic mapping $f: \mathbb{R}^q\rightarrow\mathbb{R}^D$ jointly with the latent representations $\mathbf{z}$. This is achieved by marginalizing the mapping under a Gaussian process prior \citep{gp_book}.
The generative approach allows the methods to extend beyond visualization to e.g.\ missing data imputation, data augmentation and semi-supervised tasks \citep{mattei2019miwae, urtasun2007discriminative}.

\textbf{In this paper}, we 
%argue that these defining properties are not mutually exclusive, and present a fully generative model that is both topology and distance preserving. %aims to further unify non-linear dimensionality reduction methods by using techniques from Riemannian Geometry, metric learning and 
%
learn a Riemannian manifold using Gaussian processes on which distances on the manifold match the \emph{local} distances as is implied by the Riemannian assumption.
Assuming the observed data lies on a Riemannian $q$-submanifold of $\mathbb{R}^D$ with infinite injectivity radius, then our approach can learn a $q$-dimensional representation that is isometric to the original manifold. Similar statements only hold true for traditional manifold learning methods that embed into $\mathbb{R}^q$ if the original manifold is flat. 
We learn global and local structure through a common technique from survival analysis, combined with a likelihood model based on the theory of Gaussian process arc-lengths.
Lastly, we show how the GP approach allow us to marginalize the latent representation and produce a fully Bayesian non-parametric model. 
We envision how learning probabilistic models by pairwise dissimilarities easily allow for encoding invariances.

The data handled in this paper are \emph{pairwise distances} between instances. This naturally gives a geometrical flavour to the approach since distances fall within the geometrical ontology. Note that this does not exclude tabular data --- we only require a distance can be computed between points. Further, many modern datasets come in form of pairwise distances: proteins based on their distance on a phylogenetic tree, simple GPS data for place recognition, perception data from psychology, etc.

\section{Background material}
\subsection{Gaussian Processes}
A Gaussian process (GP) \citep{gp_book} is a distribution over functions, $f:\mathbb{R}^q \rightarrow \mathbb{R}$, which satisfy that for any finite set of points $\{\mathbf{z}_i\}_{i=1}^N$, in the domain $\mathbb{R}^q$, the output $\mathbf{f}=\big(f(\mathbf{z}_1),\ldots,f(\mathbf{z}_N)\big)$ have a joint Gaussian distribution. This Gaussian is fully determined by a mean function $\mu:\mathbb{R}^q\rightarrow\mathbb{R}$ and a covariance function $k:\mathbb{R}^q\times\mathbb{R}^q\rightarrow \mathbb{R}$, such that
\begin{equation}
    p(\mathbf{f}) = \mathcal{N}(\mathbf{\mu},\mathbf{K}), 
\end{equation}
where $\mathbf{\mu}=\big(\mu(\mathbf{z}_1),\ldots,\mu(\mathbf{z}_N)\big)$ and $\mathbf{K}$ is the $N\times N$-matrix with $(i,j)$-th entry $k(\mathbf{z}_i,\mathbf{z}_j)$. 

GPs are well-suited for Bayesian non-parametric regression, since if we condition on data $\mathcal{D}=\{\mathbf{z},x\}$, where $x$ denote the labels, then the posterior of $f(\mathbf{z}^*)$, at a test location $\mathbf{z}^*,$ is given as
\begin{equation}
	p\big(f(\mathbf{z}^*)|\mathcal{D}\big)=\mathcal{N}(\mathbf{\mu}^*,\mathbf{K}^*),
\end{equation}
where
\begin{align}
    \mathbf{\mu}^*&=\mu(\mathbf{z}^*) + k(\mathbf{z}^*,\mathbf{z})^{\top}k(\mathbf{z},\mathbf{z})^{-1}x,\\
	\mathbf{K}^* &= k(\mathbf{z}^*,\mathbf{z}^*) - k(\mathbf{z}^*,\mathbf{z})^\top k(\mathbf{z},\mathbf{z})^{-1}k(\mathbf{z}^*,\mathbf{z})
\end{align}
We see that this posterior computation involves inversion of the $N\!\times\!N$-matrix $\mathbf{K}$, which has complexity $\mathcal{O}(N^3)$. To overcome this computational burden in inference we consider variational sparse GP regression, which introduces $M$ auxiliary points $\mathbf{u}$, that approximate the posterior of $f$ with a variational distribution $q$. For a review of variational GP methods, we refer to \citet{titsias-vargp}. 

\subsection{Riemannian Geometry}
A \emph{manifold} is a topological space, for which each point on it has a neighborhood that is homeomorphic to Euclidean space; that is, manifolds are locally linear spaces. Such manifolds can be embedded into spaces of higher dimension than the dimensionality of the associated Euclidean space; the manifold \emph{itself} has the same dimension as the local Euclidean space. 
A $q$-dimensional manifold $\mathcal{M}$ can, for our purposes thus, be seen as a surface embedded in $\mathbb{R}^D$. In order to make quantitative statements along the manifold we require it to be \emph{Riemannian}.
\begin{definition}
A Riemannian manifold $\mathcal{M}$ is a smooth $q$-manifold equipped with an inner product
\begin{equation}
    \langle \cdot,\cdot\rangle_{\mathbf{x}} : \mathcal{T}_{\mathbf{x}} \mathcal{M}\times\mathcal{T}_{\mathbf{x}} \mathcal{M} \rightarrow \mathbb{R}, \qquad \mathbf{x}\in\mathcal{M},
\end{equation}
that is smooth in $\mathbf{x}$. Here $\mathcal{T}_{\mathbf{x}}\mathcal{M}$ denotes the tangent space of $\mathcal{M}$ evaluated at $\mathbf{x}$.
\end{definition}
%By Nash' Embedding theorem any Riemannian $q$-manifold can be \emph{isometrically} embedded in $\mathbb{R}^D$, $D>q$. Isometric means that curves on $\mathcal{M}$ retain their length on the embedding $f(\mathcal{M})\subset\mathbb{R}^D$.

\textbf{The length of a curve}
is easily defined from the Riemannian inner product. If $\mathbf{c}:[0,1]\rightarrow \mathcal{M}$ is a smooth curve, its length is given by
%\begin{equation}
    $s = \int_0^1 \|\dot{\mathbf{c}}(t)\| \mathrm{d}t$.
%\end{equation}
On an embedded manifold $f(\mathcal{M})$ this becomes
\begin{equation}\label{curvelength}
    s = \int_0^1 \|\dot{f}(\mathbf{c}(t))\dot{\mathbf{c}}(t)\| \mathrm{d}t.
\end{equation}

A metric on $\mathcal{M}$ can then, for $ \mathbf{x},\mathbf{y}\in\mathcal{M}$, be defined as
\begin{equation}
    d_{\mathcal{M}}(\mathbf{x},\mathbf{y}) = \inf_{\mathbf{c}\in C^1(\mathcal{M})} \big\{ s | \mathbf{c}(0) = \mathbf{x} \text{ and } \mathbf{c}(1) = \mathbf{y} \big\}.
\end{equation}

\subsection{The Nakagami distribution}
We consider random manifolds immersed by a GP. The length of a curve \eqref{curvelength} on such a manifold is necessarily random as well. Fortunately, since this manifold is a Gaussian field, then curve lengths are well-approximated with the Nakagami $m$-distribution \citep{bewsher2017distribution}. %is closely related to the Gamma distribution, since we can generate Nakagami samples by transforming Gamma samples by taking their square root. 

The Nakagami distribution \citep{nakagami} describes the length of an isotropic Gaussian vector, but \citet{bewsher2017distribution} have meticulously demonstrated that this also provides a good approximation to the arc length of a GP. The Nakagami has density function
\begin{equation}
    g(s)=\frac{2m^m}{\Gamma(m)\Omega^m}s^{2m-1}\exp\Big( -\frac{m}{\Omega}s^2\Big), \qquad s\geq 0,
\end{equation}
and it is parametrised by $m\geq \sfrac{1}{2}$ and $\Omega>0$; here $\Gamma$ denotes the Gamma function. The parameters are interpretable by the equations
\begin{equation}\label{Eq:NakaParameters}
    \Omega = \mathbb{E}[s^2]\quad \text{and}\quad m = \frac{\Omega^2}{\text{Var}(s^2)},
\end{equation}
which can be used to infer the parameters through samples, although it does involve a fourth moment. 

%Combining what we know, we get that if $s$ is given as in \eqref{curvelength} and $f$ is a Gaussian process, then we will, throughout this paper, approximate $s$ with a Nakagami distribution. This approximation is studied meticulously in \citep{Brewsher}.

\section{Model and variational inference}
With prerequisites settled, we now set up a Gaussian process latent variable model that is \emph{locally} distance preserving and \emph{globally} topology preserving. Notation-wise we let $\mathcal{Z}$ denote the latent representation of a dataset $\mathcal{X}=\{\mathbf{x}_i\}_{i=1}^N$, $\mathbf{x_i}\in \mathbb{R}^D$, and let $f: \mathbf{z}\mapsto\mathbf{x}$ be the generative mapping.
%We present a model and algorithm for unsupervised learning, that learns a latent representation $\mathcal{Z}$ of a dataset $\mathcal{X}=\{\mathbf{x}_i\}_{i=1}^N$, where each $\mathbf{x_i}\in \mathbb{R}^D$. Along with this find we a generative mapping $f: \mathbf{z}\mapsto\mathbf{x}$, and marginalize the latent representation. Our model is \emph{locally} distance preserving and \emph{globally} topology preserving. To this end, we first introduce the hyperparameter $\epsilon$.
\begin{figure*}
	\centering
	\resizebox{0.8\linewidth}{!}{
	\begin{subfigure}
	\centering
	\resizebox{0.9\linewidth}{!}{%
		\begin{tikzpicture}[scale=0.7]
		\begin{scope}[every node/.style={circle,thick,draw,minimum size=0.9cm}]
		\node (z) at (0,0) {$\mathbf{z}$};
		\node (J) at (2,0)	{$\mathbf{J}$};
		\node (theta) at (4,0) {$\mathbf{\theta}$};
		\node (x) at (6,0) {$\mathbf{x}$};
		\node (eps) at (6,2) {$\epsilon$};
		\end{scope}
		
		\begin{scope}[every node/.style={rectangle,thick,draw, minimum size=0.7cm}]
		\node (u) at (2,2) {$\mathbf{u}$};
		\end{scope}
		
		\begin{scope}[>={Stealth[black]},
		every edge/.style={draw,very thick},
		every node/.style={circle}]
		\path [->] (z) edge (J);
		\path [->] (z) edge[bend right=40] (theta);
		\path [->] (J) edge (theta);
		\path [->] (theta) edge (x);
		\path [->] (u) edge (J);
		\path [->] (eps) edge (x);
		\end{scope}
		\end{tikzpicture}
	    }
	\end{subfigure}
	\begin{subfigure}
	\centering
	\includegraphics[width=\linewidth]{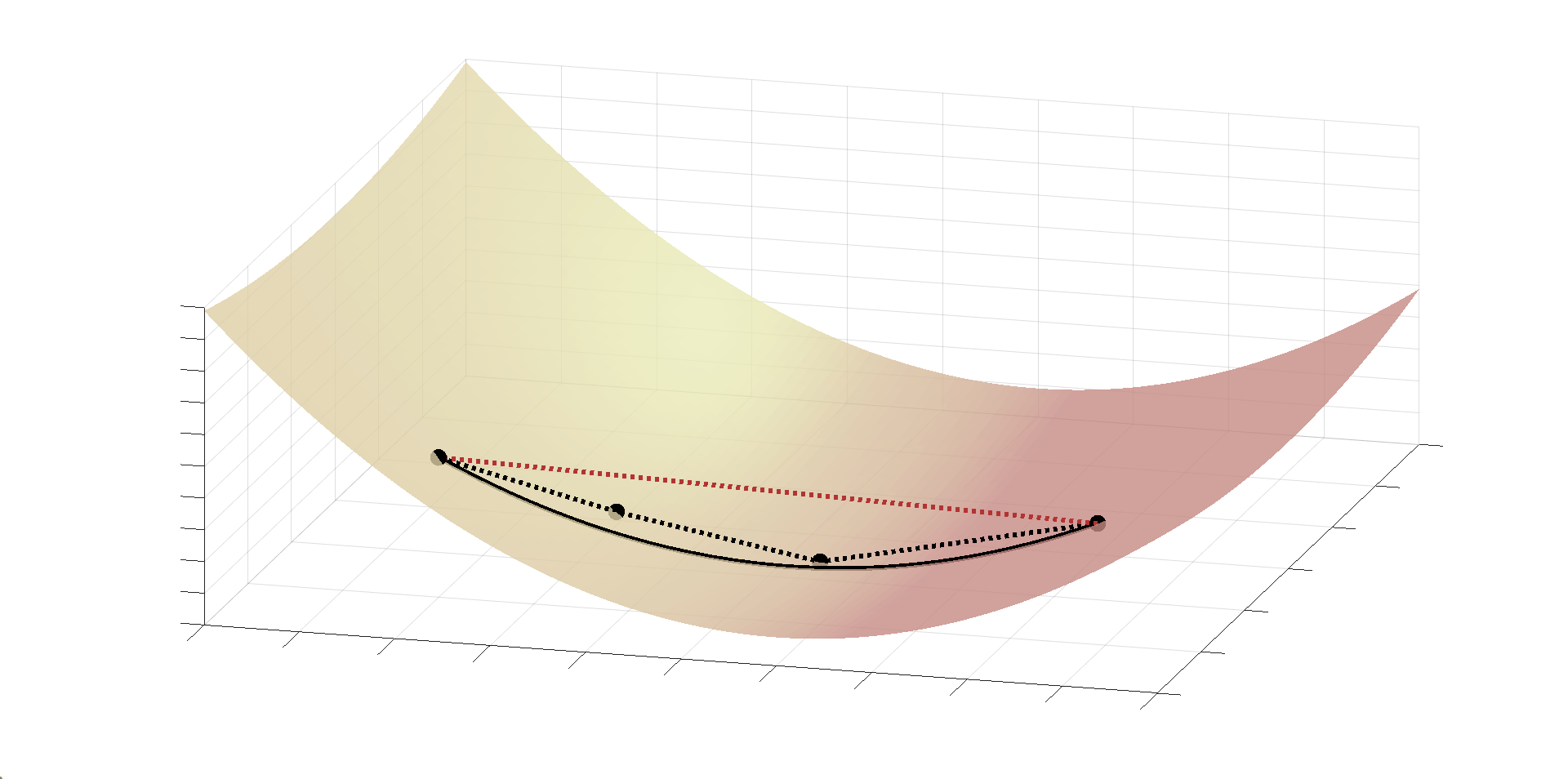}
	\end{subfigure}
	} % end of resizebox
	\caption{\emph{Left:} A graphical representation of the model: $\mathbf{x}$ is the observational input, $\mathbf{J}$ is the Gaussian process manifold and $\mathbf{\theta}$ are the parameters it yields based on latent embedding $\mathbf{z}$. $\epsilon$ is a hyperparameter for the neighbor-graph embedding and $\mathbf{u}$ are variational parameters.
	  \emph{Right:} Illustration of the task: the dashed lines are Euclidean distances in three dimensions. The black ones are \emph{neighbors} and their distance along the two-dimensional manifold should \emph{match} the 3d-Euclidean distance. The red is not a neighbor-pair and the manifold distance should not match it.}
	  %\vspace{-1mm}
\end{figure*}
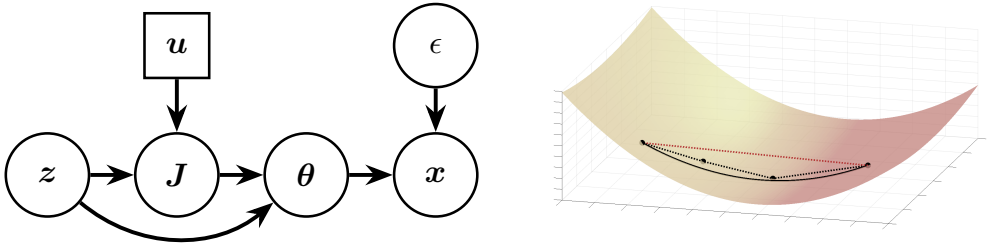

\subsection{Distance and topology preservation}  %{Graph embedding}
The \emph{manifold assumption} hypothesizes that high-dimensional data in $\mathbb{R}^D$ lie near a manifold with small intrinsic dimension. A manifold suggests that, a neighborhood around any point is approximately homeomorphic to a linear space. So nearby points are approximately linear, but non-nearby points have distances \emph{greater} than the linear approximation suggests. 

We build a Gaussian process latent variable model (GP-LVM) \citep{lawrence2005probabilistic} that is explicitly designed for distance and topology preservation.
The vanilla GP-LVM takes on the Gaussian likelihood where observations $\mathcal{X}$ are assumed i.i.d.\ when conditioned on a Gaussian process $f$. That is, $p(\mathcal{X}|f)=\prod_{i=1}^Np(\mathbf{x}_i|f(\mathbf{z}_i))$ and $p(\mathbf{x}_i|f(\mathbf{z}_i)) = \mathcal{N}(\mathbf{x}_i|f(\mathbf{z}_i),\sigma^2)$. In contrast, we consider a likelihood over pairwise distances between observations. 

\paragraph{Neighborhood graph.}
To model locality, we condition our model on a graph embedding of the observed data $\mathcal{X}$. The graph is the $\epsilon$-\emph{nearest neighbor} embedded graph; that is, the undirected graph with vertices $V=\mathcal{X}$ and edges $E=\{e_{ij}\}$, where $e_{ij}$ is in $E$, only if $d(\mathbf{x}_i,\mathbf{x}_j)< \epsilon $, for some metric $d$. Equivalently, $G=(V,E)$ can be represented by its adjacency matrix $A_G$ with entries
\begin{equation}
    a_{ij} = \mathbf{1}_{d(\mathbf{x}_i,\mathbf{x}_j)<\epsilon}.
\end{equation}
In Sec.~\ref{subsec:TDA} we discuss how to choose $\epsilon$ informedly, but for now we view it as a hyperparameter.
%Instead we will shortly discuss what it embodies. 

%That is, with respect to some distance function $d(\cdot,\cdot)$, we consider the dataset $\mathcal{E} = \{e_{ij}\}$, where $e_{ij} = d(\mathbf{x}_i,\mathbf{x}_j)$. 

\paragraph{Manifold distances.}
To arrive at a likelihood over pairwise distances, we first recall that the linear interpolation between $\mathbf{z}_i$ and $\mathbf{z}_j$ in the latent space has curve length
\begin{equation}\label{Eq:ManifoldLength}
    s_{ij} = \int_0^1 \|\mathbf{J}(\mathbf{c}(t))\dot{\mathbf{c}}(t)\| \mathrm{d}t, \quad \mathbf{c}(t) = \mathbf{z}_i(1-t)+\mathbf{z}_jt,
\end{equation}
where $\mathbf{J}$ denotes the Jacobian of $f$, which is our generative manifold approximation.

As the manifold distance $d_{\mathcal{M}}$ is the length of the shortest connecting curve, then $s_{ij}$ is by definition an upper bound on $d_{\mathcal{M}}$. However, as the manifold is locally homeomorphic to a Euclidean space, then we can expect $s_{ij}$ to be a good approximation of the distance to nearby points, i.e.
\begin{align}
  d_{\mathcal{M}}(\mathbf{z}_i, \mathbf{z}_j) &\approx s_{ij} \qquad \text{for } \| \mathbf{x}_i - \mathbf{x}_j \| < \epsilon \\
  d_{\mathcal{M}}(\mathbf{z}_i, \mathbf{z}_j) &\leq s_{ij}    \qquad \text{otherwise.}
\end{align}
The behavior we seek is that local interpolation in latent space should mimic local interpolation in data space only if the points are close in data space. If they are far apart, they should \emph{repel} each other in the sense that the linear interpolation in latent space should have \emph{large} curve length.

\paragraph{Censoring.} To encode this behavior in the likelihood, we introduce \emph{censoring} \citep{lee2003statistical} into our objective function. This method is usually applied to missing data in survival analysis, when the event of something happening is known to occur later than some time point.
%This is known as \emph{right-censoring}, the opposite -- the event happened before some time-point, is called \emph{left-censoring}. 

We may think of censoring as modeling inequalities in data. The censored likelihood function for i.i.d.\ data $t_i$ following distribution function $G_\theta$, with density function $g_\theta$, is defined
\begin{equation}
    L(\{t_i\}_{i=1}^N|\theta,T)=\prod_{t_i<T}g_\theta(t_i)\prod_{t_i\geq T}(1-G_\theta(T)),
\end{equation}
where $\theta$ are the parameters of the distribution $G$ and $T$ is some `time point', where the experiment ended.
\citet{carreira2010elastic} remark that most neighborhood-embedding methods have loss functions with two terms: one attracting close point and one scattering term for far away connections.
Censoring provides a \emph{likelihood} with similar such terms. It may be viewed as a probabilistic version of the ideas in \emph{maximum variance unfolding} \citep{weinberger2006introduction}.

%\paragraph{Loss function.} 
%Our objective is to find a representation that preserves the geometric features of data. To this end we shall not consider the observation $\{\mathbf{x}_i\}_{i=1}^N$ themselves, but rather the pairwise distances they constitute. That is, with respect to some distance function $d(\cdot,\cdot)$, we consider the dataset $\mathcal{E} = \{e_{ij}\}$, where $e_{ij} = d(\mathbf{x}_i,\mathbf{x}_j)$. 

%The manifold assumption is that locally the distances between points is close to the Euclidean, while points that are not close must be further apart, on the manifold, than the distance of the embedding space: the Euclidean. Hence, we apply censoring on the Euclidean distances from the embedding: we trust small distances to be good representatives of the manifold distance, but large distances can not be represented by Euclideans.
%We now lean on censoring to encode the neighborhood graph. Conceptually, we only trust distance observations between points that are sufficiently close in the embedding space, i.e.\ points that are connected by an edge in the graph.
%The manifold assumption is that locally the distances between points is close to the Euclidean, while points that are not close must be further apart, on the manifold, than the distance of the embedding space: the Euclidean. Hence, we apply censoring on the Euclidean distances from the embedding: we trust small distances to be good representatives of the manifold distance, but large distances can not be represented by Euclideans.
\begin{figure*}
\begin{align}\label{Eq:loglikelihood}
    l\left(\!\big\{\{e_{ij}\}_{i < j} \big\}_{i=1}^{N-1} \Big | \theta,\epsilon\right)
      =& -\!\!\sum_{e_{ij}<\epsilon}\! \left(\log\Gamma\left(m_{ij}\right) \!+\!
          m_{ij}\log\left(\frac{\Omega_{ij}}{m_{ij}}\right) \!-\!
          (2m_{ij} \!-\! 1) \log\left(e_{ij}\right) \!+\!
          \frac{m_{ij}e_{ij}^2}{\Omega_{ij}}
          \right)\nonumber\\
       & -\!\!\sum_{e_{ij}\geq \epsilon}\! \left( \log\Gamma\left(m_{ij}\right) \!-\!
          \log\left(\Gamma\left(m_{ij}\right) \!-\!
          \gamma(m_{ij}, \frac{m_{ij}}{\Omega_{ij}}e_{ij}^2)\right) \right),
\end{align}
\vspace{-4mm}
\caption{The likelihood of our model. Here $\Gamma$ and $\gamma$ denotes the Gamma function and lower incomplete gamma function respectively and $m_{ij}$ and $\Omega_{ij}$ are the Nakagami-parameters of Eq.~\ref{Eq:ManifoldLength}.}
\label{fig:likelihood}
\end{figure*}

\paragraph{Local distance likelihood.}
From earlier, we know that if the manifold $f(\mathcal{M})$ is a Gaussian field, then distances \eqref{Eq:ManifoldLength} are approximately Nakagami distributed. Thus, we write our likelihood as  %function
\begin{equation*}\label{likelihood}
L(\big\{\{e_{ij}\}_{i < j} \big\}_{i=1}^{N-1}|\theta,\epsilon)=\prod_{e_{ij}<\epsilon}g_\theta(e_{ij})\prod_{e_{ij}\geq \epsilon}(1-G_\theta(\epsilon)),
\end{equation*}
where $G_\theta$ is the distribution function of a Nakagami with parameters $\theta=\{m,\Omega\}$.
The resulting log-likelihood is given in Eq.~\ref{Eq:loglikelihood} within Fig.~\ref{fig:likelihood}.

%Hence, the log-likelihood we shall maximise is
%\begin{align}\label{Eq:loglikelihood}
%    l\left(\!\big\{\{e_{ij}\}_{i < j} \big\}_{i=1}^{N-1} \Big | \theta,\epsilon\right)
%      =& -\!\!\sum_{e_{ij}<\epsilon}\! \left(\log\Gamma\left(m_{ij}\right) \!+\!
%          m_{ij}\log\left(\frac{\Omega_{ij}}{m_{ij}}\right) \!-\!
%          (2m_{ij} \!-\! 1) \log\left(e_{ij}\right) \!+\!
%          \frac{m_{ij}e_{ij}^2}{\Omega_{ij}}
%          \right)\nonumber\\
%       & -\!\!\sum_{e_{ij}\geq \epsilon}\! \left( \log\Gamma\left(m_{ij}\right) \!-\!
%          \log\left(\Gamma\left(m_{ij}\right) \!-\!
%          \gamma(m_{ij}, \frac{m_{ij}}{\Omega_{ij}}e_{ij}^2)\right) \right),
%\end{align}
%where $\Gamma$ and $\gamma$ denotes the Gamma function and lower incomplete gamma function respectively and $m_{ij}$ and $\Omega_{ij}$ are the Nakagami-parameters of Eq.~\ref{Eq:ManifoldLength}.

Until now, we have introduced the log-likelihood based on an $\epsilon$-NN graph, that preserves geometric features. Next we marginalize all other parameters to make a Bayesian model.

\subsection{Marginalizing the representation}
We have a loss function \eqref{Eq:loglikelihood} that matches distances $e_{ij}$ with parameters $\theta_{ij}=\{m_{ij},\Omega_{ij}\}$. We now seek to first fit these parameters and marginalize them to obtain a full Bayesian approach. First, we will assume that conditioned on $\theta$, we get the independent observations, i.e. 
\begin{align}
    p(\mathcal{E}|\theta,\epsilon)
      &= \prod_{1\leq i<j\leq N}p(e_{ij}|\theta_{ij},\epsilon) \\
      &= L\left(\big\{\{e_{ij}\}_{i < j} \big\}_{i=1}^{N-1}|\theta,\epsilon\right),
\end{align}
as known from Eq.~\ref{likelihood}. We infer these parameters of the Nakagami by introducing a latent Gaussian field $J$ and a latent representation $\mathbf{z}$. This allows us to define curve length \eqref{Eq:ManifoldLength}, which we assume is also Nakagami distributed. In practice, we draw\footnote{We can approximate $s$ by finely discretizing $\mathbf{c}$ and sum over the integrand.} $m$ samples of $s_{ij}$ from Eq.~\ref{Eq:ManifoldLength}, and estimate the mean and variance of their second moment. This gives estimates of $m_{ij}$ and $\Omega_{ij}$ via Eq.~\ref{Eq:NakaParameters}.

Essentially, we match distances on the manifold $J$ with the observed distances $\mathcal{E}$. We marginalize this manifold
\begin{align}
    p(\mathcal{E}|\mathbf{z})&=\int p(\mathcal{E}|\theta)p(\theta|\mathbf{J},\mathbf{z})p(\mathbf{J}) \mathrm{d}\theta \mathrm{d}\mathbf{J},\label{Eq:cond_liklihood} %\\
    %\text{where}\quad 
\end{align}
where
\begin{align}
    p(\theta|\mathbf{J},\mathbf{z})&:=\int p(\theta|s)p(s|\mathbf{J},\mathbf{z})\mathrm{d}s, \\\text{and}\quad p(\theta|s)&=\begin{cases}\delta_{\mathbb{E}s^2}(\Omega)\\
    \delta_{\nicefrac{\Omega}{\text{Var}(s^2)}}\big(m\big),\end{cases}
\end{align}
and $\delta$ denotes the Dirac probability measure and $p(s|\mathbf{J},\mathbf{z})$ is the approximate Nakagami distribution  \eqref{Eq:ManifoldLength}. This means that $s_{ij}$ and $e_{ij}$ are both Nakagami variables that share the same parameters, which interpretively means the manifold distances $s_{ij}$ match the embedding distances $e_{ij}$.

Further, we can pose a prior on $\mathbf{z}$ and marginalize this in Eq.~\ref{Eq:cond_liklihood}. We infer everything variationally \citep{blei2017variational}, and choose a variational distribution over the marginalized variables. We approximate the posterior $p(\theta,\!\mathbf{J},\!\mathbf{z},\!\mathbf{u}|\mathcal{E})$ with\looseness=-1
\begin{equation}
    q(\theta,\mathbf{J},\mathbf{z},\mathbf{u}):=q(\theta|\mathbf{J},\mathbf{z})q(\mathbf{J},\mathbf{u})q(\mathbf{z}),
\end{equation}
where $\mathbf{u}$ is an inducing variable \citep{titsias-vargp}, and
\begin{align}
    q(\theta|\mathbf{J},\mathbf{z})=p(\theta|\mathbf{J},\mathbf{z}),\quad q(\mathbf{J},\mathbf{u})=p(\mathbf{J}|\mathbf{u})q(\mathbf{u})\\
    \text{and}\quad q(\mathbf{z})=\mathcal{N}(\mathbf{\mu}_z,\mathbf{A}_z),
\end{align}
where $\mathbf{\mu}_z$ is a vector of size $N$ and $\mathbf{A}_z$ is a diagonal $N\!\times\!N$-matrix. Further $q(\mathbf{u})=\mathcal{N}(\mathbf{\mu}_u,\mathbf{S})$ is a full $M$-dimensional Gaussian.

This allow us to bound the log-likelihood \eqref{Eq:loglikelihood}, with the evidence lower bound (ELBO)
\begin{align}
    \log p(\mathcal{E})
      &=\log\int \frac{p(\mathcal{E},\theta,\mathbf{J},\mathbf{z},\mathbf{u})}{q(\theta,\mathbf{J},\mathbf{z},\mathbf{u})}q(\theta,\mathbf{J},\mathbf{z},\mathbf{u})\mathrm{d}\theta \mathrm{d}\mathbf{J}\mathrm{d}\mathbf{u}\mathrm{d}\mathbf{z}\\
% Samme ligning som expectation
%      &=\log\mathbb{E}_{q(\theta,\mathbf{J},\mathbf{z},\mathbf{u})} \left[
%          \frac{p(\mathcal{E},\theta,\mathbf{J},\mathbf{z},\mathbf{u})}{q(\theta,\mathbf{J},\mathbf{z},\mathbf{u})}
%        \right]\\
    %&=\log\int p(\mathcal{E}|\theta)\frac{p(\mathbf{J}|\mathbf{z}}{}
    \begin{split}
    &\geq \mathbb{E}_\theta[l(\mathcal{E}|\theta)] - \text{KL}\big(q(\mathbf{u})||p(\mathbf{u})\big) \\
    &\hspace{19.5mm} - \text{KL}\big(q(\mathbf{z})||p(\mathbf{z})\big),\label{ELBO}    
    \end{split}
\end{align}
where both KL-terms are analytically tractable, but the first term has to be approximated using Monte Carlo. The right hand side here is readily optimized with gradient descent type algorithms.

In summary, we have a latent representation $\mathcal{Z}$ and a Riemannian manifold immersed as a GP $\mathbf{J}$. This implies that between any two points $\mathbf{z}_i$ and $\mathbf{z}_j$, we can compute $s_{ij}$, which is approximately Nakagami. With censoring we can match $s_{ij}$ with observation $e_{ij}$, if $e_{ij}<\epsilon$; else we push $s_{ij}$ to have all its mass on $[\epsilon,\infty)$. It is optimized with variational inference by maximizing Eq.~\ref{ELBO}.

\subsection{Invariances and geometric constraints}
\emph{Why} is it worth learning the manifold in a coordinate-free way? Invariances are easily encoded via dissimilarity pairs by introducing equivalence classes in saying $d(\mathbf{x}_i,\mathbf{x}_j)=0$ if $\mathbf{x}_i$ and $\mathbf{x}_j$ are in the same equivalence class. Popular choices of such equivalence classes are rotations, translations and scaling. Many constraints one could wish to impose on models can be formulated as geometric constraints. It holds true also for GPLVM-based models as seen in \citet{urtasun2008topologically}, who wish to encode topological information, and \citet{zhang2010invariant}, who highlight invariant models' usefulness in causal inference.
Geometric constraints can alternatively be encoded with GPs that take their output directly on a Riemannian manifold \citep{mallasto:2018}. \citet{pmlr-v119-kato20a} try to enforce geometric constraints in Euclidean autoencoders by changing the optimisation, and \citet{miolane2020learning} build Riemannian VAEs.

The geometry of latent variable models in general is an active field of study \citep{arvanitidis2018latent, Tosi:UAI:2014}, and \citet{simard2012transformation} and \citet{kumar2017semi} argues that the tangent (Jacobian) space serves a convenient way to encode invariances. Recently, \citet{borovitskiy2020matern} developed a framework for GPs defined on Riemannian manifold. Contrary to their method, we learn the manifold where they a priori determine it.

\subsection{Topological Data Analysis and the influence of $\epsilon$}\label{subsec:TDA}
The model is naturally affected by the hyperparameter $\epsilon$. We argue that it can be chosen in a geometrically founded way using Topological Data Analysis \citep{carlsson2009topology}. By constructing a \emph{Rips diagram} \citep{fasy2014} one can find $\epsilon$ such that the $\epsilon$-NN graph captures the right topology of data. It is beyond this paper to summarize the techniques; we refer readers to \citet{chazal2017introduction}.

To understand what $\epsilon$ means in broader terms we can study corner cases. If $\epsilon\!=\!\infty$ we would match \emph{all} observed distances, which resembles MDS. If the covariance function of the marginalized $\mathbf{J}$ is constant\footnote{In this case the generating function $f$ has a linear kernel.} the latent space is also preserved (scaled) Euclidean, hence iso-GPLVM may in this setting be viewed as a probabilistic MDS. This links well with how the GPLVM generalized the probabilistic PCA \citep{lawrence2005probabilistic}.\looseness=-1

Although we shall not further discuss it in this paper, the Bayesian setup also suggests $\epsilon$ could potentially be marginalized. The argument why this is not as straightforward as one could hope is that the model has a pathological solution in the corner case $\epsilon \!=\! 0$. In this case, all points would repel each other, and a high likelihood can be obtained without a meaningful representation.

\section{Experiments}
We perform experiments first on a classical toy dataset and on the image datasets COIL20 and MNIST. We refer to the presented model as \emph{Isometric Gaussian Process Latent Variable Model} (Iso-GPLVM). For comparisons we evaluate other models also based on dissimilarity data. In all cases we initialize Iso-GPLVM with IsoMap, as it is known GP-based methods are sensitive to initialization \citep{init-gplvm}. We use the Adam-optimizer \citep{kingma2014adam} with a learning rate of $3\cdot10^{-3}$ and optimize sequentially $q(\mathbf{z})$ and $q(\mathbf{u})$ separately. We use $m=100$ inducing points for $q(\mathbf{u})$ and an ARD-kernel as covariance function.

\subsection{Swiss roll}
\begin{figure*}
    \centering
    \includegraphics[width=\textwidth]{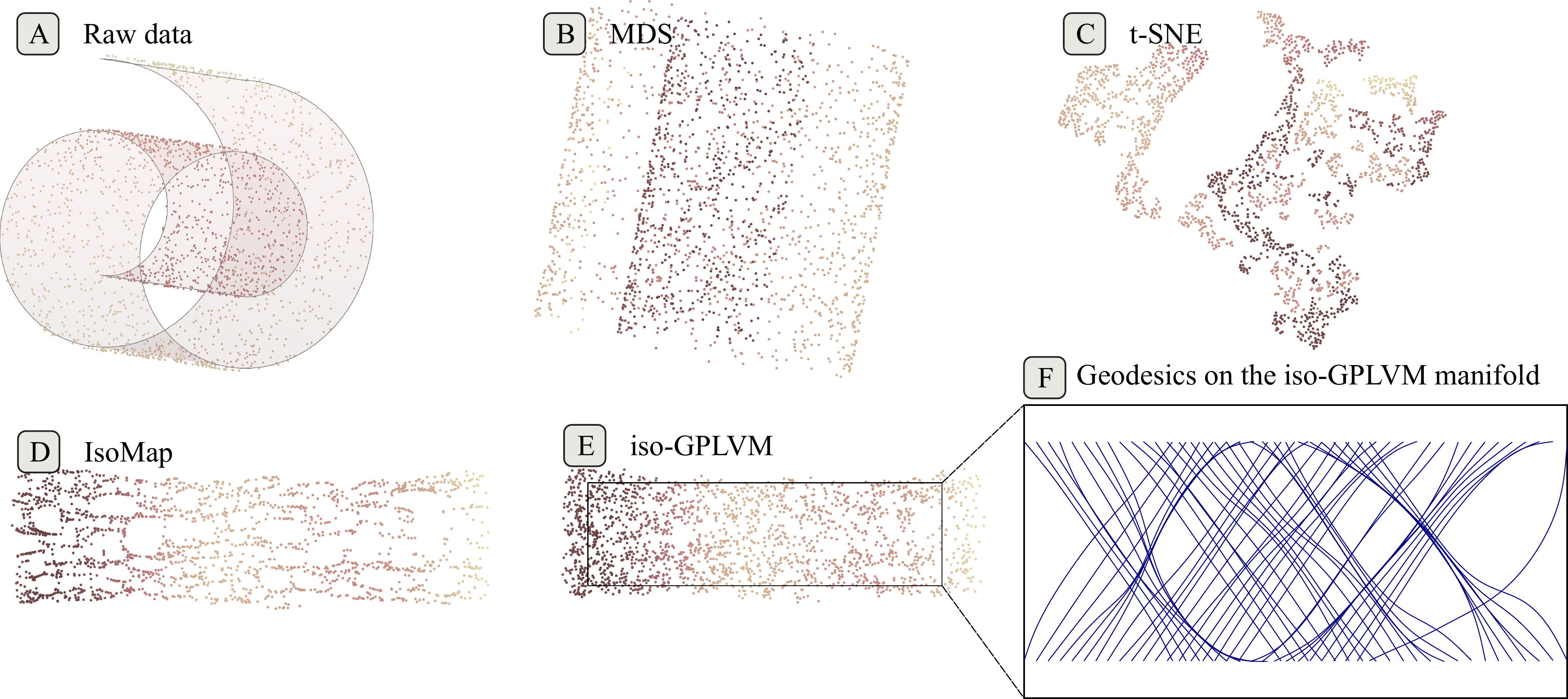}
    \vspace{-6mm}
    \caption{Data (A) and embeddings (B--E). All embeddings are shown
      with a unit aspect ratio to highlight that only IsoMap (D) and Iso-GPLVM (E) recover
      the elongated structure of the swiss roll. (F) shows some geodesics on the learned 2-dimensional manifold.}
    \label{fig:swiss}
    %\vspace{-1mm}
\end{figure*}

The `swiss roll' was introduced by \citet{tenenbaum:global:2000} to highlight the difficulties of non-linear manifold learning. The point cloud resides on a 2-dimensional manifold embedded in $\mathbb{R}^3$ and can be thought as a paper rolled around itself (see Fig~\ref{fig:swiss}A).
%\begin{figure}
%\centering
%\begin{subfigure}{0.2\textwidth}
%\includegraphics[width = \hsize]{plots/3dswiss_edit2.png}
%\caption{The dataset in 3D.}
%\label{swissroll-data}
%\end{subfigure}
%\begin{subfigure}{0.77\textwidth}
%\includegraphics[width = \hsize]{plots/latents_swiss.pdf}
%\caption{\textit{Top left}: MDS, \textit{top right}: IsoMap, \textit{bottom left}: t-SNE, \textit{bottom right}: Iso-GPLVM.}
%\label{swiss:embeddings}
%\end{subfigure}
%    \caption{The Swiss roll dataset and different latent embeddings. The color of the points on the right indicate the distance to the center of the swirl.}
%    \label{fig:swissroll}
%\end{figure}

We find a $2$-dimensional latent embedding by four methods: MDS, t-SNE, IsoMap and Iso-GPLVM. From Fig.~\ref{fig:swiss} we observe the linear MDS is unable to capture the highly non-linear manifold. t-SNE captures some local structure, but the global outlook is far from the ground truth. We tried several tunings of the perplexity hyperparameter (60 in the plot), none successfully captured the structure. It is known that t-SNE is prone to create clusters, even if clusters are not a natural part of a dataset \citep{amid2018}.\looseness=-1

Naturally, as the dataset was constructed for the `geodesic' approach of IsoMap, this captures both global and local structure. On closer inspection, we see the linear interpolations, stemming from Dijkstra's algorithm, leaves some artificial `holes' in the manifold. Hence, on a smaller scale it can be argued the topology of the manifold is captured imperfectly. The plot suggests Iso-GPLVM closes these holes and approximates the topology of an unfolded paper.%, which was supported by a Rips diagram, as described in Section \ref{subsec:TDA}.
\begin{figure}[t]
    \centering
    \includegraphics[width = 0.9\hsize]{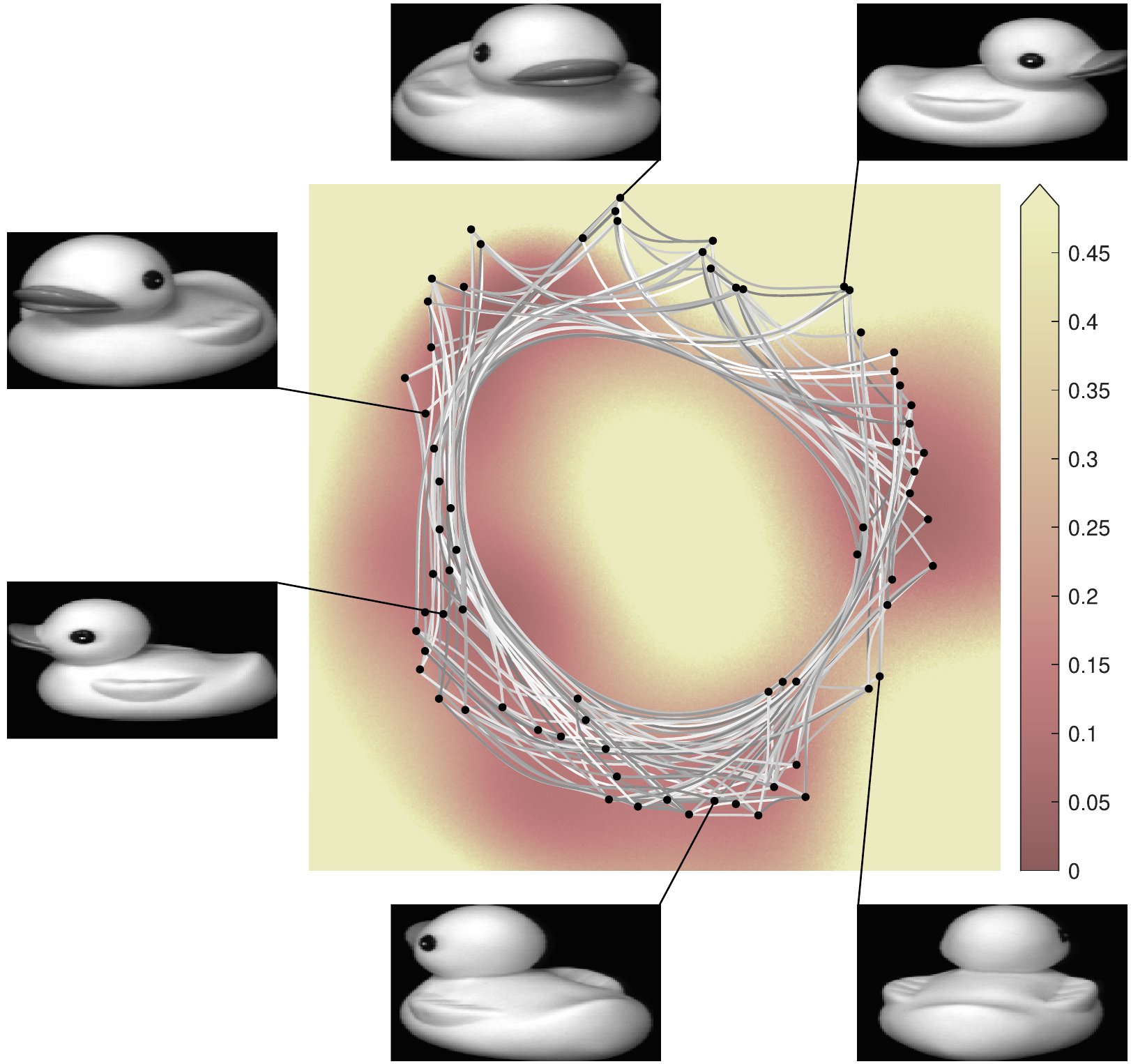}
    \vspace{-1mm}
    \caption{The 2-dimensional embeddings of the 72 images of a rubber duck. We observe from the geodesics (grey curves) how the latent manifold has learned the circular nature of the data.}
    \label{fig:rubberduck}
    \vspace{-3mm}
\end{figure}

Figure~\ref{fig:swiss}F visualizes some geodesics and they appear roughly linear. There is some 'gathering` fix points which are due to the sparsity of the GP. These geodesics inform us that not only is the representation good, but the learned \emph{geometry} is correct since the geodesics match those we know from Fig.~\ref{fig:swiss}A. We used $\epsilon=0.4$.

\subsection{COIL20}
COIL20 \citep{nene} consists of greyscale images of 20 objects photographed from 72 different angles spanning a full rotation (see Figure \ref{fig:rubberduck} for some examples). This implies in total 1440 images --- the version we use is of size $128\times 128$ pixels, thus the original data resides in $\mathbb{R}^{16384}$.

\begin{figure*}[t!]
    \centering
    \includegraphics[width = \hsize]{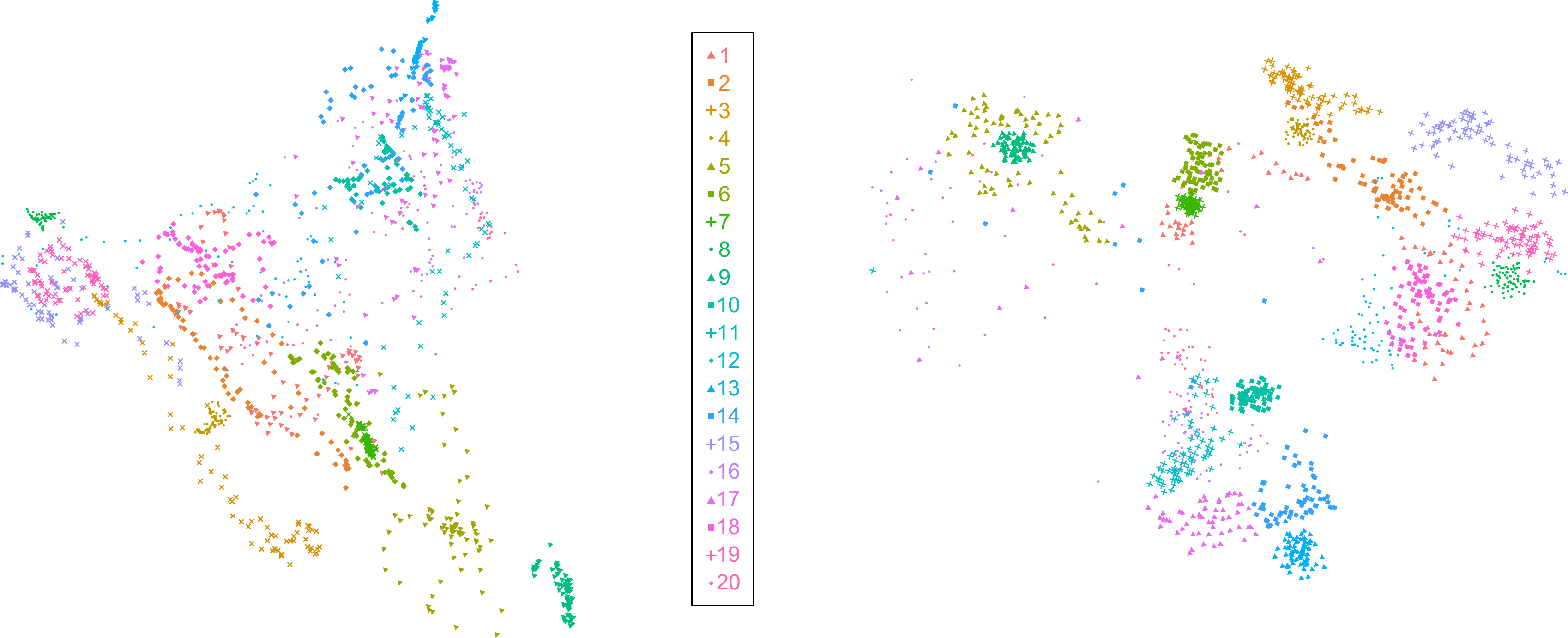}
    \vspace{-3mm}
    \caption{Embeddings of COIL20 objects. \emph{Left}: IsoMap and \emph{right}: Iso-GPLVM. We see that globally Iso-GPLVM can separate the objects (color- and shape coded), but is not able to find to all local structures.}
    \label{fig:COIL20-embeddings}
    \vspace{-2mm}
\end{figure*}

First, we focus on only one object --- a rotated rubber duck --- to highlight the geodesic behaviour. Figure~\ref{fig:rubberduck} shows the 2-dimensional embeddings and the geodesic curves on the learned manifold in latent space. We clearly observe the circular structure we expect from the rotated duck. On top of this the geodesics show the Riemannian geometry of the latent space: they move along the data manifold and avoid the space where no data is observed.
The background color is the measure $\mathbb{E}\big[$\scalebox{0.7}{$\sqrt{\det (\mathbf{J}^\top\!\mathbf{J})}$}$\big]$,
which provide a view of the Riemannian geometry of the latent space.
\citet{bishop1997magnification} call this measure the \emph{magnification factor}.
Large values (light color) imply trajectories moving in this area are longer
and likely also more uncertain \citep{hauberg2018only}.

IsoMap, t-SNE, UMAP and others, are also able to infer the circular embedding, but Iso-GPLVM is the only model to infer a \emph{geometry} on latent space. For IsoMap the latent geometry is implicitly Euclidean through it's loss \eqref{eq:stress}, and t-SNE and UMAP do not allow for geodesic computations.\looseness=-1

When considering all 20 objects at once a global element of separating the distinct objects is a key task to infer the topological structure. The embeddings for IsoMap and Iso-GPLVM are visible in Fig.~\ref{fig:COIL20-embeddings}. Here IsoMap struggle to clearly separate objects due to it's implicit assumption of one connected manifold.  Iso-GPLVM finds the \emph{global} topological structure, but in no instances finds the local structure. So why is it unsuccessful here when successful in Fig.~\ref{fig:rubberduck}?  When considering all $1440$ images we only use $100$ inducing points, and in this view it is unsurprising that the model has to use most capacity on the global structure. In Fig.~\ref{fig:rubberduck} there is no sparsity required since there is only $72$ images, and there is enough capacity to detect the hole in the manifold. This is a common problem for GP-based methods.

\subsection{MNIST}
\paragraph{Metrics.} We evaluate our model on 5000 images from MNIST, and we foremost wish to highlight how invariances can be encoded with dissimilarity data. We consider fitting our model to data under three different distance measures. We consider the classical Euclidean distance measure
\begin{equation}
    d(\mathbf{x}_i,\mathbf{x}_j) = \| \mathbf{x}_i - \mathbf{x}_j \|.
\end{equation}
Further, we consider a metric that is invariant under image rotations
\begin{equation}
    d_{\textsc{ROT}}(\mathbf{x}_i,\mathbf{x}_j) = \inf_{\theta \in [0, 2\pi)}\Big\{ d\left( R_{\theta}(\mathbf{x}_i), \mathbf{x}_j \right) \Big\},
\end{equation}
where $R_{\theta}$ rotates an image by $\theta$ radians. We note $d_{\textsc{ROT}}(\mathbf{x}_i,\mathbf{x}_j)\leq d(\mathbf{x}_i,\mathbf{x}_j)$ always.
Finally, we introduce a \emph{lexicographic} metric \citep{rodriguezvelazquez2018lexicographic}
\begin{equation}
d_{\textsc{LEX}}(\mathbf{x}_i,\mathbf{x}_j) = \begin{cases} \epsilon, &\hspace{-2mm}\text{if }y_i \neq y_j\\
\min\{2r,\ d(\mathbf{x}_i,\mathbf{x}_j)\}, &\hspace{-2mm}\text{if }y_i = y_j\end{cases}
\end{equation}
which in the censoring phase enforce images carrying different labels to repel each other. This is a handy way to encode a topology or clustering based on discrete variables, when such are available. For all metrics, we have normalized the data and have set $\epsilon=7$.
\begin{figure*}
    \centering
    \includegraphics[width = \hsize]{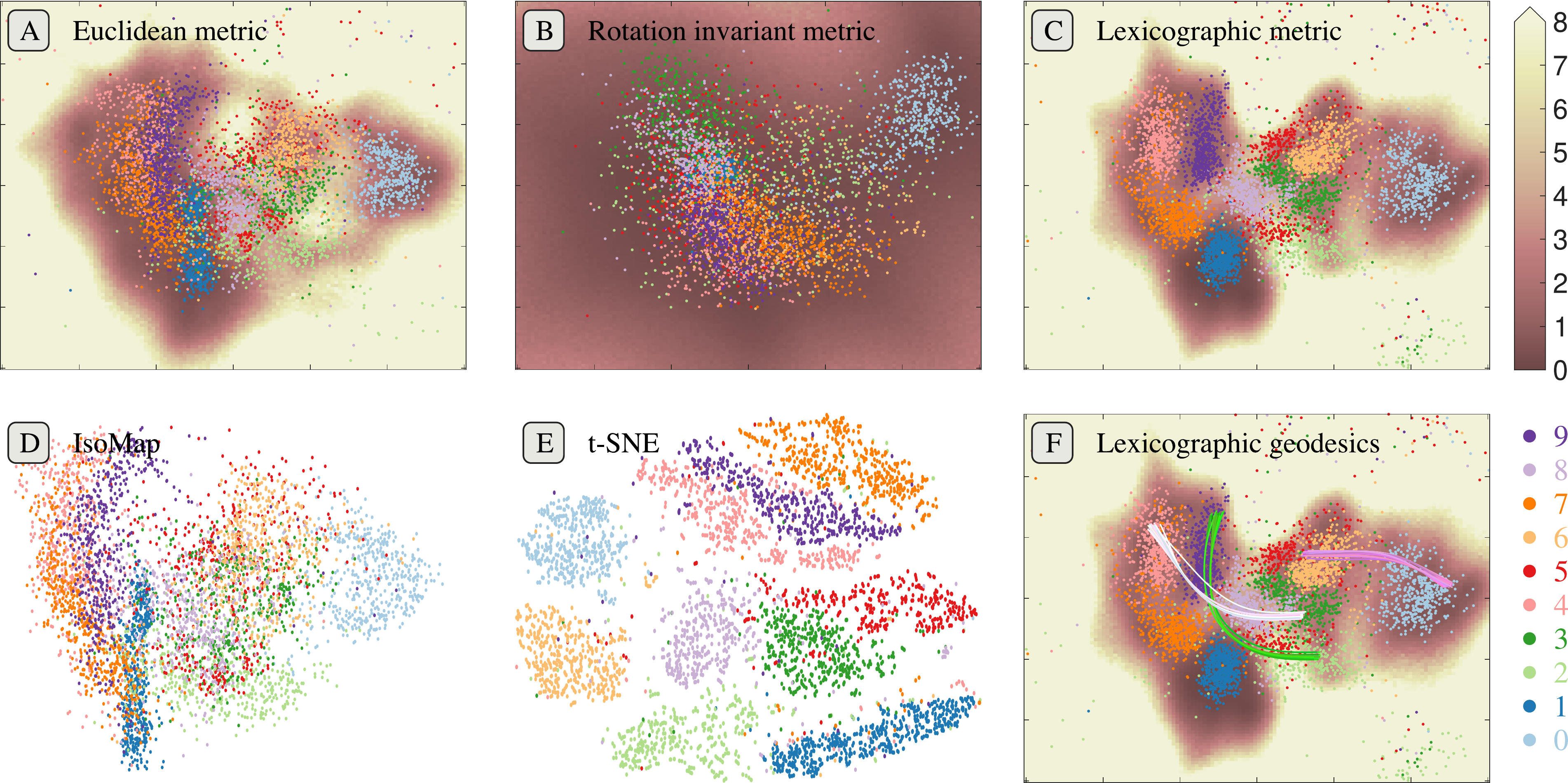}
    \caption{Embeddings of MNIST attained with our method under different
      metrics (A---C) and for baselines IsoMap (D) and t-SNE (E).
      The background color show the expected volume measure associated
      with the Riemannian metric $\mathbb{E}\big[$\protect\scalebox{0.7}{$\sqrt{\det (\mathbf{J}^\top\!\mathbf{J})}$}$\big]$.
      A large measure generally indicate high uncertainty of the manifold.
      Panel F shows Riemannian geodesics under the lexicographic metric.}
    \label{fig:MNIST-embeddings}
    %\vspace{-1mm}
\end{figure*}

% (a full white color is a value larger than 6). It is clear that when the topology is encoded with the lexicographic metric, a much clearer separation of the integers is notable. 

\paragraph{Results.} Figure~\ref{fig:MNIST-embeddings}(A---C) show the latent embeddings of the three metrics. The background color again indicates the magnification factor $\mathbb{E}\big[$\scalebox{0.7}{$\sqrt{\det (\mathbf{J}^\top\!\mathbf{J})}$}$\big]$.
Panels A, D and E base their latent embedding on the Euclidean metric. We observe that IsoMap (D) and Iso-GPLVM (A) appear similar in shape, unsurprisingly as we initialize with IsoMap, but Iso-GPLVM finds a cleaner separation of the digits. Particularly, this is evident for the \emph{six, three and eight digits}. The \emph{fives} seem to group into several tighter cluster, and this behavior is found for t-SNE as well. Overall, from a clustering perspective, t-SNE 
visually is superior; but distances \emph{between} clusters in (A) can be larger than the straight lines that connect them. This is evident from the lighter background color between cluster, say, \emph{zeros} and \emph{threes}. We note that IsoMap and t-SNE has no associated Riemannian metric and as such distances between any input cannot be computed.\looseness=-1

The rotation invariant metric results in a latent embedding where different classes significantly overlap.
Upon closer inspection we, however, note several interesting properties of the embedding.
\emph{Zero digits} are well separated from other classes as a rotated 0 does not resemble
any other digits; the \emph{one digits} form a cluster that is significantly more compact
than other digits as there is limited variation left after rotations have been factored out;
\emph{two and five digits} significantly overlap, which is most likely due to 5 digits resembling
2 digits when rotated $180^{\circ}$; similar observations hold for the \emph{four, nine and
six digits}; and a partial overlap between \emph{three and eight digits} as is often observed.
The overall darker background is due to the rotational invariant metric being shorter than
the Euclidean counterpart.

In terms of clustering the lexicographic approach outshines the other metrics. This is expected
as the metric use label information, but neatly illustrate how domain-specific
metrics can be developed from weak or partial information. Most classes are well-separated
except for a region in the middle of the plot. Note how this region has high uncertainty.

The Riemannian geometry of the latent space implies that geodesics (shortest paths) can be computed in our model. Figure~\ref{fig:MNIST-embeddings}F shows example geodesics under the lexicographic metric. Their highly non-linear appearance emphasizes the curvature of the learned manifold. The green geodesics has one endpoint in a cluster of nine digits and move along this cluster avoiding the uncertain area of eights and fives, as opposed to linearly interpolating through them.

\section{Discussion}
We introduced a model for non-linear dimensionality reduction from dissimilarity data. It is the first of its kind based on Gaussian processes. The non-linearity of the method stems both from the Gaussian processes, but also from the censoring in the likelihood. It unifies ideas from Gaussian processes, Riemannian geometry and neighborhood graph embeddings. Unlike traditional manifold learning methods that embed into $\mathbb{R}^q$, we embed into a $q$-dimensional Riemannian manifold through the learned metric. This allows us to learn latent representations that are isometric to the true underlying manifold.\looseness=-1

The model does have limitations.
Aesthetically the visualizations are not as satisfactory as e.g. t-SNE. However, the access to a geometrically founded GPLVM is of interest to many practitioners, since GPs are ubiquitous in many sub-disciplines of machine learning such as Bayesian optimization and reinforcement learning. Here, GPs are fundamental parts of decision-making pipelines, whereas t-SNE is a valuable visualization technique. 
%The existence of the embeddings are guaranteed by Nash's Embedding theorem \citep{nash-riemann}.
The Nakagami distribution that approximates the arc lengths of Gaussian processes is prone to overestimate the variance \citep{bewsher2017distribution} and better approximations would improve our method. Further, the model inherits problems of optimizing the latent variables and it has previously been noted that good performance in this regime is linked with good initialization \citep{init-gplvm}. 

Our experiments highlight that
Iso-GPLVM can learn the geometry of data and geometric constraints are 
easier encoded by learning a manifold contra doing GP regression. The uncertainty quantification associated with GPs follow through and further highlights the 
connection between uncertainty, geometry and topology. To the best of our knowledge, our model is the first of its kind that, locally, can asses the quality of the manifold approximation through the associated Riemannian measure.

%\subsection{Generating new samples}
%All inference thus far has been done in a \emph{coordinate-free} manner; in other words, we have yet to embed our manifold $f(\mathcal{M})$ in $\mathbb{R}^D$. We can do this embedding with Euclidean isometries, translation and rotation, and inspired by the fundamental theorem of analysis
%\begin{align}
%	f(\mathbf{z}_i) & = f(\mathbf{z}_j) + \int_0^1 j(\mathbf{c}(t))\dot{\mathbf{c}}(t)\mathrm{d}t, \\
%	\mathbf{c}(t) &= \mathbf{z}_j(1-t) + \mathbf{z}_i t.
%\end{align}
%In this view, the translation part can be done by the original points, as we assume $f(\mathbf{z}_j)\approx \mathbf{x}_j$ and we can, for a new point $\mathbf{z}^*$, define a generator as
%\begin{equation}
%	x^* := f(\mathbf{z}_i) + \mathfrak{R}(\mathbf{z}_i)\int_{0}^{1}\mathbf{J}(\mathbf{c}(t))\dot{\mathbf{c}}(t)\mathrm{d}t,
%\end{equation}
%with $\mathbf{c}(t)=\mathbf{z}_i(1-t) + \mathbf{z}^* t$, and $\mathfrak{R}$ is a $D\times D$ rotation-matrix, that can be optimized to best fit with the original data $\mathcal{D}$ and $\mathbf{J}$ is the inferred Jacobian from Eq.~\ref{ELBO}. This is a rather naive way, since it needs many local embeddings and follows the intuition of \citet{local-tangents}. A more principled way would be to learn an isometry $f$ by regression methods.

\section*{Acknowledgements}
This project has received funding from the European Research Council (ERC)
under the European Union’s Horizon 2020 research and innovation programme (grant agreement no 757360). MJ and SH were supported in part by a research grant (15334) from VILLUM FONDEN. MJ is supported by the Carlsberg Foundation (CF20-0370). The majority of this work was done while MJ was affiliated with the Technical University of Denmark.

\bibliographystyle{icml2021}
\bibliography{doc-iso}

\end{document}